\pdfoutput=1
\documentclass[11pt]{article}

\usepackage{EACL2023}

\usepackage{times}
\usepackage{latexsym}

\usepackage{multirow}
\usepackage{multicol}
\usepackage{graphicx}
\usepackage[shortlabels]{enumitem}
\usepackage{xcolor}
\usepackage{float}
\usepackage{listings}
\usepackage{booktabs}
\usepackage{hyperref}

\usepackage[T1]{fontenc}
\usepackage[utf8]{inputenc}

\usepackage{microtype}

\usepackage{inconsolata}

\title{EventNet-ITA: Italian Frame Parsing for Events}

\author{Marco Rovera \\
  Fondazione Bruno Kessler, Trento, Italy \\
  \texttt{m.rovera@fbk.eu}
  }

\begin{document}
\maketitle
\begin{abstract}
This paper introduces EventNet-ITA, a large, multi-domain corpus annotated \textit{full-text} with event frames for Italian. Moreover, we present and thoroughly evaluate an efficient multi-label sequence labeling approach for Frame Parsing. Covering a wide range of individual, social and historical phenomena, with more than 53,000 annotated sentences and over 200 modeled frames, EventNet-ITA constitutes the first systematic attempt to provide the Italian language with a publicly available resource for Frame Parsing of events, useful for a broad spectrum of research and application tasks. Our approach achieves a promising 0.9 strict F1-score for frame classification and 0.72 for frame element classification, on top of minimizing computational requirements. The annotated corpus and the frame parsing model are released under open license.
\end{abstract}

\section{Introduction}
\label{sec:intro}
Frame Parsing is a powerful tool for real-world applications in that it enables deep grasp of the meaning of a textual statement and automatic extraction of complex semantic descriptions of situations, including events, and their relations with the entities involved. To this effect, Frame Parsing can effectively be used for Event Extraction, as the two tasks share the common goal of recognizing and classifying argument structures of a target predicate. However, training supervised models for Frame Parsing is a data-intensive task, which is why comprehensive linguistic resources are available in few languages, thereby limiting further research and application in downstream tasks.
Furthermore, most existing corpora are created by targeting the annotation of one single frame (or event) class per sentence. While lexicographically motivated, this procedure makes the training of automatic models and their application to real-world scenarios more complicated and burdensome. 

The contribution described in this paper is twofold: first, we present EventNet-ITA (EvN-ITA) a large-scale, multi-domain corpus annotated \textit{full-text} (see Section \ref{sec:annotation}) with over 200 semantic frames of events \cite{fillmore2001frame} and 3,600 specific frame elements in Italian, also discussing the motivation behind its creation, the annotation guidelines and the covered domains; secondly, we introduce an efficient multi-label sequential approach for eventive Frame Parsing and evaluate it on the dataset.
This work aims at providing the community with a solid, manually-curated corpus and a ready-to-use model for frame-based event extraction for Italian, thus filling an existing data gap. In fact, recent works in application fields like computational social science \citep{minnema2021frame, minnema2022sociofillmore} or historical NLP \citep{sprugnoli2017one, menini2023scent} showed how semantic frames can be used as a powerful textual analysis tool to investigate a wide range of societal and historical phenomena. The annotated dataset, along with its full documentation, is released to the community under open license (see Section \ref{sec:release}). The envisioned application purpose of EvN-ITA is that of enabling accurate mining of events from large collections of documents, with focus on individual, social and, in a broad sense, historical~phenomena. \\
The paper is structured as follows: Section \ref{sec:related} discusses existing work in Frame Parsing and Event Extraction, with a subsection focused on Italian; Section \ref{sec:dataset} introduces our annotated corpus and describes the motivations and design decisions that guided its creation, while Section \ref{sec:annotation} focuses on the annotation procedure. In Section \ref{sec:methodology} we discuss the methodology for Frame Parsing, a transformer-based multi-label sequence labeling approach. In Section \ref{sec:results} we evaluate our methodology and discuss the results. Section \ref{sec:release} provides the reader with pointers for the dataset and model release, while Section \ref{sec:conclusions} concludes the paper and highlights future directions of our work.

\section{Related Work}
\label{sec:related}

The development of systems able to recognize and classify event mentions and their argument structure in text has been a long-term effort in computational linguistics and a variety of methods has been employed for the task of Event Extraction \citep{ahn2006stages,liao2010using,chen2015event,nguyen2016joint,orr-etal-2018-event,nguyen2019one,lu2021text2event, paolini2021structured}.
Event Extraction is the task of recognizing and classifying event mentions and entities involved in the event from a textual statement and it has seen applications in a wide range of fields, like social media analysis \cite{de2019global}, biomedical NLP \citep{li2019biomedical,huang2020biomedical,ramponi2020biomedical}, history and humanities \citep{segers2011hacking,cybulska2011historical,sprugnoli2019novel,lai-etal-2021-event, rovera2021event}, as well as literary text mining \cite{sims-etal-2019-literary}.
Although benchmark datasets exist, like Automatic Content Extraction (ACE) \cite{walker2006ace} for Event Extraction or TAC-KBP \cite{ellis2015overview} for multiple event-related tasks, they exhibit limitations in terms of size and domain coverage. Also, while they are well suited for evaluation campaigns, they have not been designed for use in real-world application tasks. Moreover, most of these corpora only exist for English, with few extensions for other languages \cite{ji2016overview}.

\subsection{Frame Parsing}
Frame Parsing \citep{das2014frame,swayamdipta2017frame,swayamdipta2018syntactic} consists in recognizing, in a textual expression, a word or set of words (the \textit{lexical unit}) as the predicate evoking a given frame and isolating the text spans that evoke the semantic arguments (\textit{frame elements}) related to that frame. Frames are conceptual structures describing prototypical situations and their realizations in text. The reference linguistic resource for Frame Parsing in English is FrameNet (FN) \citep{baker1998berkeley, fillmore2001frame,ruppenhofer2006framenet}. In this work, we use Frame Parsing for extracting event frames. While event extraction initiatives have been based on a variety of models, approaches and schemes, which are not always interoperable or comparable, the advantage of using Frame Parsing for Event Extraction is the availability of an established resource, based on a unified, grounded theoretical framework \cite{fillmore1976frame}. EvN-ITA differs from FN in that the latter is based on lexicographic annotation (one target lexical unit per sentence), providing only a small subset of full-text annotated data \citep{ruppenhofer2016framenet}, whereas EvN-ITA has been annotated by design in a \textit{full-text} fashion (see Section \ref{sec:annotation}). Also, it is important to point out that EvN-ITA is not meant to be a comprehensive Italian version of the popular English FN. Instead, in this work we adopt part of the FN schema but focus exclusively on event-denoting frames, aiming at providing a large, self-contained and robust tool for frame-based Event Extraction in Italian.

\subsection{Italian Event Extraction and Frame Semantics}
As for Italian, the Frame Labeling
over Italian Texts Task (FLAIT) was organized at EVALITA in 2011 \cite{basili2012evalita}. Moreover, Event Extraction in Italian was the object of the EVENTI evaluation campaign at EVALITA 2014 \cite{caselli2014eventi}, which focused on temporal processing and was based on the Ita-TimeBank schema \cite{caselli2011annotating}. Later on, \citet{caselli2018italian} experimented with the same dataset using a neural architecture and evaluated the impact of different word embeddings for event extraction. While Italian Event Extraction approaches have traditionally been based on the TimeML \cite{sauri2006timeml} classification scheme, which provides 7 broad, temporal-oriented classes, more recently the necessity has emerged of a more fine-grained annotation schema for event classification, as discussed by \citet{sprugnoli2017one}. Supported by a survey involving historians, the authors investigated the application of event extraction on historical texts. \citet{sprugnoli2019novel} describe a specific schema, adapting semantic categories provided by the Historical Thesaurus of the Oxford English Dictionary (HTOED) \cite{kay2009historical}, resulting in 22 topic-driven event classes, thereby moving towards developing a richer and at the same time finer-grained inventory of classes for representing events in text. 
As for frame semantics, on the other hand, \citet{basili2017developing} and \citet{brambilla2020automatic} described a work in progress for the creation of IFrameNet, a large scale Italian version of FN, by using semi-automatic methods and manual validation for frame induction, with 5,208 sentences annotated with at least one lexical unit. However, the dataset has not been released so far. In fact, despite the considerable amount of work in lexical \cite{lenci2012lexit,jezek2014t} and frame semantics \cite{tonelli2008frame, tonelli2009semi, lenci2010building, lenci2012enriching}, Italian still lacks an extensive, publicly available linguistic resource for Frame Parsing. 

\section{Dataset}
\label{sec:dataset}

\subsection{EventNet-ITA}
\label{sec:eventnet}
In order to ensure multilingual compatibility, we employ a selection of event frames from FN \cite{baker1998berkeley,ruppenhofer2006framenet}, where available. This way, 85\% of EvN-ITA classes are mapped to FN schema, either by direct match (59\%) or by subclassing (26\%). In a minor number of cases (15\% of the schema), where target phenomena are not covered in FN, an \textit{ad hoc} frame class has been created. Frame-to-frame mappings between EvN-ITA and FN are provided in the documentation of the resource. Table \ref{tab:enetstats} offers a quantitative description of the corpus.
\begin{table}[h]
\centering
\resizebox{1\linewidth}{!}{
\begin{tabular}{lr}
Annotated sentences & 53,854 \\
Tokens & 1,583,612 \\
Vocabulary (words) & 97,512\\
Avg. sentence length (tokens) & 29 \\
Modeled event frames & 205 \\
Modeled frame elements & 3,571 \\
Lexical units & 837 \\
Frame instances & 102,294 \\
Frame element instances & 180,279 \\
Frame instances per sentence (avg.) & 1.9 \\
Examples per class (avg.) & 491 \\
\end{tabular}}
\caption{\label{tab:enetstats}
Statistics of the EvN-ITA dataset.}
\end{table}
EvN-ITA counts 53,854 annotated sentences, including negative examples (see Section \ref{sec:guidelines}), 102,294 event instances and over 1.5 million tokens, annotated by an experienced annotator (native speaker) with background in Frame Semantics. Each frame class has on average 491 annotated examples.\\ 
The corpus - as well as the annotation schema - has been created with the purpose of covering historical narratives in a broad sense, but without committing to a specific textual genre. For this reason, as well as for creating a releasable corpus, sentences for the annotation set of EvN-ITA have been sampled from a subset of the Italian Wikipedia edition. In order to filter out irrelevant documents, i.e. documents not likely to contain events, we collected Wikipedia pages falling under the categories \textit{Events by country}\footnote{\url{https://it.wikipedia.org/wiki/Categoria:Eventi_per_stato}} and \textit{History by country}.\footnote{\url{https://it.wikipedia.org/wiki/Categoria:Storia_per_stato}} This choice ensures a wide variety of featured events, both temporally (from ancient history to the present days) and geographically.\\
Through standard pre-processing (tokenization, lemmatization and dependency parsing have been performed using TINT\footnote{\url{https://dh.fbk.eu/research/tint/}} \cite{palmero2018tint}), a pool of sentences, arranged by lemma, was generated, from which to pick for the annotation set. Annotated sentences are drawn from 16,309 different Wikipedia articles.

\subsection{Domain coverage}
\begin{figure}
    \centering
    \includegraphics[width=0.49\textwidth]{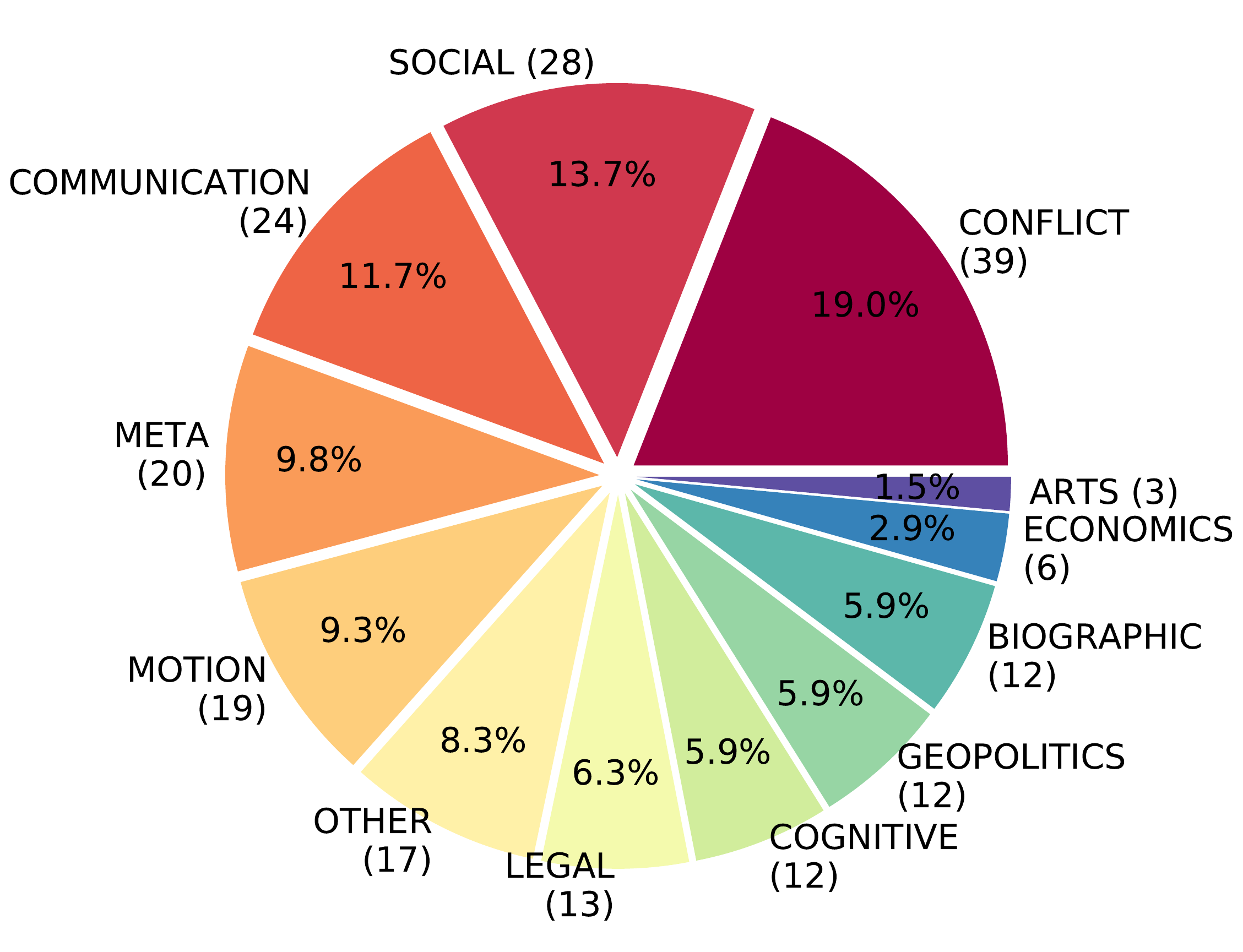}
\caption{Macro-topics covered in EvN-ITA (in brackets, the number of frames belonging to each domain).}
    \label{fig:macroclasses}
\end{figure}

In the design phase of the resource, a manual analysis was made of existing corpora in multiple languages, in order to circumscribe the domains and classes to be modelled. Resources as Automatic Content Extraction \cite{doddington2004automatic, linguistic2005ace}, Event Nugget \cite{mitamura2015event}, the Historical Thesaurus of the Oxford English Dictionary (HTOED) \cite{kay2009historical} and FN \cite{baker1998berkeley} were reviewed and compared. FN is currently the most complete, rich and established existing resource and has been taken as reference for the development of EvN-ITA. This choice is motivated by the opportunities it offers in terms of reuse, coverage and possible multilingual extensions.
EvN-ITA's annotation schema covers 205 different event frames, each provided with a set of specific frame elements (unique modeled frame elements amount to 3,571), and has been extensively documented by providing, for each frame, its definition, the corresponding set of lexical units and frame elements associated to it. The distribution of classes, arranged by topic, is depicted in Figure \ref{fig:macroclasses}.
%\begin{table}[h]
%\centering
%\begin{tabular}{lc}
%\hline
%\textbf{Topic} & \textbf{Classes}\\
%\hline
%CONFLICT & 39 \\
%SOCIAL & 28 \\
%COMMUNICATION & 24 \\
%META/FUNCTIONAL & 22 \\
%MOTION & 19 \\
%OTHER & 17 \\
%LEGAL/JURIDICAL & 13 \\
%BIOGRAPHIC & 11 \\
%COGNITIVE & 12 \\
%GEOPOLITICS & 12 \\
%ECONOMICS & 6 \\
%ARTS & 3 \\
%\hline
%\end{tabular}
%\caption{\label{tab:macroclasses}
%Macro-topics currently covered in EvN-ITA, along with the number of fine-grained event frames belonging to each domain.}
%\end{table}
Beside conflict-related events, that hold a prominent place in historical accounts and journalistic narratives, we have taken care to extend the collection of event types to other aspects of the life of societies and individuals, such as legislative and legal processes, work, establishment of and membership in social organizations, life events, as well as events related to the arts, economic processes and cognitive processes such as decisions, skills, judgements, amongst others. In the design of the resource, attention has been paid also to maintain the balance between internal coherence and usability of the class schema. This has been achieved in multiple ways:
\begin{enumerate}[(a)]
    
    \item by including for each class, where existing, also its opposite (\textsc{Hiring} / \textsc{Firing}, \textsc{Create Social Entity} / \textsc{End Social Entity}, \textsc{Vehicle Take Off} / \textsc{Vehicle Landing});

    \item by making possible narrative chains (e.g. \textsc{Committing Crime}, \textsc{Arrest}, \textsc{Trial}, \textsc{Sentencing}, \textsc{Imprisonment}, \textsc{Captivity}, \textsc{Releasing});

    \item by providing couples of classes representing an event and the subsequent logical state (e.g. \textsc{Becoming a Member} / \textsc{Membership}, \textsc{Becoming Aware} / \textsc{Awareness}, \textsc{Make Acquaintance} / \textsc{Acquaintance}).

    \item by ensuring a certain degree of redundancy and perspective (\textsc{Being In Place} / \textsc{Temporary Stay}, \textsc{Being Born} / \textsc{Giving Birth}). 
    
\end{enumerate}

\section{Annotation}
\label{sec:annotation}

\begin{figure*}[ht]
\includegraphics[width=1.0\textwidth]{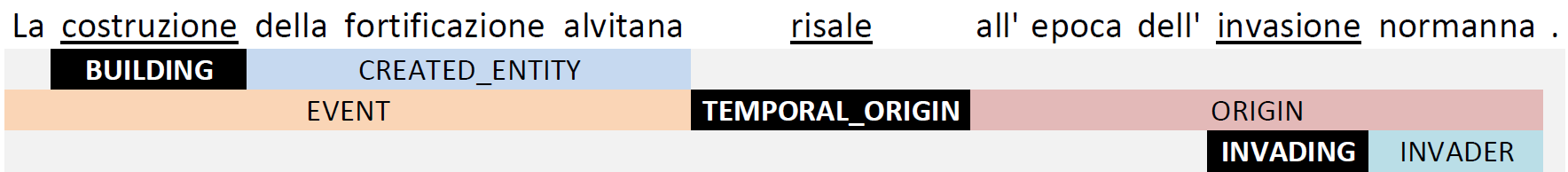}
\caption{An example of full-text annotation in EvN-ITA (English translation: \textit{The construction of the Alvitian fortification dates back to the time of the Norman invasion.)}.}
\label{fig:annotationexample}
\end{figure*}

The textual corpus, generated as discussed in Section \ref{sec:eventnet}, has been manually annotated by labeling event triggers with their frame class and predicate arguments with the corresponding frame element. Annotation was performed at the sentence level and was conducted frame-driven, by first selecting significant event frames for the domain and subsequently identifying the most relevant lexical units for each frame. Given a sentence, \textit{any} lexical units in our schema and all related frame elements are annotated, producing as many layers as there are event mentions (\textit{full-text} annotation). Figure \ref{fig:annotationexample} shows an example of full-text annotation from EvN-ITA.
\subsection{Format}
The IOB2 annotation format is being used, in which the B-tag identifies the first token of a span, the I-tag identifies all tokens inside the span and the O-tag all out-of-mention tokens. Discontinuous mentions are allowed, both for frames and for frame elements. The only constraint for event mentions is that they cannot overlap: each token in a sentence can denote at most one event type. This does not hold for frame elements: in fact, given a sentence with multiple frame occurrences, frame elements from different annotation sequences (i.e. belonging to different frames) can always overlap, hence a token may be labeled with more than one frame element tag\footnote{In addition, a frame element may overlap with a frame mention if they belong to different annotation sequences, which is quite often the case. Only an frame-frame overlap is excluded.} (See Figure \ref{fig:annotationexample}).
\subsection{Guidelines}
\label{sec:guidelines}
EvN-ITA is thoroughly documented, both in the form of general annotation guidelines (what to annotate) and at the annotation schema level (frame description, lexical units, frame elements).\\
As for lexical units, we exclusively focus on nouns, verbs and multi-word expressions. Although also other parts-of-speech (adverbs, for example) can be loosely event-evoking, this focus is motivated by practical reasons: nouns and verbs, along with multi-word expressions, are the most frequent triggers of event mentions in text and are characterized by a richer syntactic structure, which in turn is crucial for harvesting information related to frame elements. Nouns are annotated as event triggers only if they reference \textit{directly} the occurrence of an event, but not when the reference is indirect, for example:
\begin{quote}
[...] \textit{culminarono a Blois nel 1171 con la [morte \textbf{\textsc{Death}}] sul rogo di 31 ebrei}. (culminated in Blois in 1711 with the death at the stake of 31 Jews.)
\end{quote}
\begin{quote}
\textit{Nell'aprile del 1700, Giovanni si ammalò terribilmente e si trovò quasi sul letto di [morte \textbf{\textsc{Ø}}]}. (In April 1700, John fell terribly ill and was nearly on his deathbed.)
\end{quote}
As for verbs, we annotate the main verb but not the auxiliary.
\begin{quote}
\textit{Appena ricostruita dalla devastazione del sisma la città fu [distrutta \textbf{\textsc{Destroying}}] nuovamente} [...] (As soon as it was rebuilt from the devastation of the earthquake, the city was destroyed again [...])
\end{quote}
Multi-word expressions are annotated when they break compositionality, for example in \textit{radere al suolo} (raze to the ground) or \textit{aprire il fuoco} (opening fire), or in verbal periphrastic use \textit{essere al corrente} (being aware) or \textit{fare visita} (paying a visit).\\
Frame mentions are annotated regardless of their factuality value, which means that also negated or hypotetical frame mentions must be annotated, as well as those introduced by modals. Conversely, in EvN-ITA we do not annotate as frame mention lexical units that are used with metaphorical meaning or in the form of rhetoric expression.\footnote{More annotation examples for such cases are provided in Appendix \ref{appendix:appendixa}.}\\
EvN-ITA's annotation schema consists of 205 frames and 837 lexical units, out of which 358 have at least 100 annotations each, 191 have a number of annotations comprised between 50 and 99, and 288 have a number of annotations comprised between 20 an 40. The annotation process has been oriented to keep the balance between frame completion and polysemy preservation. For this reason, we also annotated less frequent lexical units encountered in the corpus, resulting in a long queue of lexical units with less than 20 occurrences each. This strategy was adopted in order both to increase the flexibility of the resource and to set the stage for its future extension.  
Also, with the aim of improving robustness, for each lexical unit we annotated a number of \textit{negative examples}, i.e. sentences in which the given lexical unit occurs without triggering any of the corresponding frames.\footnote{See Appendix \ref{appendix:appendixa} for negative examples.}\\ 
Within the scope of this work, we consider as events any accomplishment, achievement or process, without distinction. The schema additionally models a number of states (e.g. \textsc{Being  In Place}, \textsc{Captivity}) and relations (\textsc{Leadership}, \textsc{Duration Relation}, \textsc{Possession}). 
As for semantic roles, we referred to FN's frame elements, with minor adaptations or additions, which in most cases tend towards increased specificity.

\subsection{Inter-Annotator Agreement}  
EvN-ITA was annotated by one single native speaker annotator with a solid background in Frame Semantics. For this reason, particular attention has been devoted to assessing the robustness, consistency and intelligibility of the resource by means of inter-annotator agreement analysis. We therefore validated our schema and guidelines by re-annotating 2,251 sentences, spanning over 61 classes, with a second (native speaker) annotator.
When selecting classes to be included in this validation set, we paid attention to include pairs or triplets of frames with high semantic similarity,\footnote{For example: \textsc{Blaming} / \textsc{Accuse}, \textsc{Reporting} / \textsc{Denounce}, \textsc{Marriage} / \textsc{Being\_Married}.} in order to stress the test. We used two different metrics for assessing agreement: Jaccard Index (computed as the ratio between the number of items annotated with the same label and the sum of all annotated items) and Cohen's Kappa \cite{cohen1960coefficient}. The scores have been computed both at token level (relaxed) and at span level (strict). Results are reported in Table \ref{tab:agreement}.
\begin{table}[h!]
    \centering
    \resizebox{1\linewidth}{!}{
    \begin{tabular}{lll|ll}
    \hline
     & \multicolumn{2}{c|}{\textbf{Jaccard}}  & \multicolumn{2}{c}{\textbf{Cohen's K}}  \\
     \hline
     & Token & Span & Token & Span  \\
    \hline
    Lexical units & 0.952 & 0.945 & 0.951 & 0.944 \\
    Frame elements & 0.878 & 0.832 & 0.877 & 0.830 \\
    \end{tabular}
    }
    \caption{Inter-annotator agreement scores.}
    \label{tab:agreement}
\end{table}
Considering the high number of different frames and frame elements in EvN-ITA, we observe that agreement values are high and indicate that the guidelines are sufficiently detailed in their description of the linguistic phenomena to be annotated. As expected, the annotation of frame elements has proven more challenging. A further manual analysis conducted on a sample of 245 sentence pairs with low agreement showed that disagreement had three main sources:
\begin{description}
    \itemsep0em
    \item[Ontological] (67,5\%) a textual span is recognized as frame or frame element by one annotator but not by the other;
    \item[Span Length] (20,4\%) annotators agree on the label but not on the exact span to annotate;
    \item[Classification] (12\%) annotators agree on the span to annotate but assign two different labels.
\end{description}

\section{Methodology}
\label{sec:methodology}
A traditional approach for Frame Parsing, as well as for Event Extraction, is to break down the problem into sub-tasks \citep{das2014frame,ahn2006stages}, usually separating the steps of trigger identification, frame classification and argument extraction. However, a major downside of this approach, besides being more complex, is that it implies error propagation from higher-level sub-tasks downwards. Instead, we propose to learn all the tasks in one single step, allowing the model to simultaneously exploit tag relations on the time (sequence) axis and on the token axis. Thus, in this work Frame Parsing is approached end-to-end and is treated as a multi-label sequence tagging problem. The strength of this design option lies in its simplicity as it requires minimal pre-processing and does not imply the use of additional knowledge, as well as in its efficiency, as it minimizes computational requirements (see~Section~\ref{sec:experimental}).
\subsection{Preprocessing}
\label{sec:preprocessing}
The adoption of full-text annotation implies, at preprocessing time, the definition of each frame element as frame-specific, in order to avoid overlaps between frame elements with the same name but referring to different frames. In fact, many frames belonging to the same semantic area share a set of frame elements with the same name. For example, both motion frames \textsc{Fleeing} and \textsc{Motion\_Downwards} have a frame element called \textsc{Mover}. In EvN-ITA, frame elements referring to the same \textit{semantic role} (thus carrying the same name) but belonging to different frames are assigned different, frame-aware labels. Therefore, frame elements \textsc{Mover-Fleeing} and \textsc{Mover-Motion\_Downwards} will be assigned two different labels. This data encoding strategy, in return, allows us to minimize the need for post-processing (as each predicted frame element is implicitly linked to its frame) and enables the model to learn relationships between multiple frame elements occurring on the same token/span.

\subsection{Experimental Setup}
\label{sec:experimental}
Our frame parser aims at jointly extracting all frame mentions and all related frame elements in a target sentence. In other words, given an input sentence, each token must be labeled according to the event frame and/or frame element(s) it denotes. %We could not find, so far, any evidence in the literature for this method being tested on an extensive dataset for Frame Parsing. 
The underlying idea is to leverage mutual co-occurrence between frame (and frame element) classes, as certain frames typically tend to appear more often with, or have a semantic preference for, other frames.\footnote{This assumption is mentioned in previous work \cite{liao2010using} and we verified it in our dataset by analysing frame relationships
with several co-occurrence measures, such as Pointwise Mutual Information (see
Appendix \ref{appendix:appendixb}).}
This way, the model is led to not only learn correspondences between a word and a given frame or frame element, but also local patterns of co-occurrence between different frame elements.\\
\begin{table*}[ht!]
    \centering
    \begin{tabular}{ll|ccc|ccc}
    \toprule
    & & \multicolumn{3}{c|}{\textbf{Frames}} & \multicolumn{3}{c}{\textbf{Frame Elements}} \\
    & & \multicolumn{3}{c|}{\textit{n} = 40} & \multicolumn{3}{c}{\textit{n} = 200} \\
    \cmidrule(r){3-5}\cmidrule(r){6-8}
    & & P & R & F1 & P & R & F1\\
    \cmidrule(r) {3-5}\cmidrule(r){6-8}
    \multirow{6}{2.5em}{TEST} & All classes & 0.904 & 0.914 & \textbf{0.907} & 0.841 & 0.724 & \textbf{0.761} \\
    & All classes weighted & 0.909 & 0.919 & \textbf{0.913} & 0.85 & 0.779 & \textbf{0.804} \\
    & Best \textit{n} classes & 0.974 & 0.982 & 0.978 & 0.938 & 0.912 & 0.923 \\
    & Worst \textit{n} classes & 0.811 & 0.808 & 0.806 & 0.72 & 0.441 & 0.516 \\
    & \textit{n} most frequent classes & 0.912 & 0.933 & 0.922 & 0.861 & 0.831 & 0.843 \\
    & \textit{n} least frequent classes & 0.865 & 0.871 & 0.865 & 0.781 & 0.493 & 0.575 \\
    \bottomrule
    \end{tabular}
    \caption{\textit{Token-based} (\textit{relaxed}) performance for multi-label sequential Frame Parsing (macro average, aggregate). Figures in bold represent the reference performance values for EventNet-ITA.}
    \label{tab:reseventstoken}
\end{table*}
\begin{table*}[ht!]
    \centering
    \begin{tabular}{ll|ccc|ccc}
    \toprule
    & & \multicolumn{3}{c|}{\textbf{Frames}} & \multicolumn{3}{c}{\textbf{Frame Elements}} \\
    & & \multicolumn{3}{c|}{\textit{n} = 40} & \multicolumn{3}{c}{\textit{n} = 200} \\
    \cmidrule(r) {3-5}\cmidrule(r){6-8}
    & & P & R & F1 & P & R & F1\\\cmidrule(r) {3-5}\cmidrule(r){6-8}
    \multirow{6}{2.5em}{TEST} & All classes & 0.906 & 0.899 & \textbf{0.901} & 0.829 & 0.666 & \textbf{0.724} \\
    & All classes (weighted) & 0.909 & 0.903 & \textbf{0.905} & 0.853 & 0.711 & \textbf{0.768} \\
    & Best \textit{n} classes & 0.975 & 0.976 & 0.975 & 0.937 & 0.867 & 0.898 \\
    & Worst \textit{n} classes & 0.81 & 0.789 & 0.796 & 0.673 & 0.398 & 0.476 \\
    & \textit{n} most frequent classes & 0.915 & 0.917 & 0.915 & 0.878 & 0.762 & 0.813 \\
    & \textit{n} least frequent classes & 0.879 & 0.866 & 0.87 & 0.743 & 0.441 & 0.529\\
    \bottomrule
    \end{tabular}
    \caption{\textit{Span-based} (\textit{strict}) performance for multi-label sequential Frame Parsing (macro average, aggregate). Figures in bold represent the reference performance values for EventNet-ITA.}
    \label{tab:reseventsspan}
\end{table*}
In order to provide a reliable performance assessment, we opted for an 80/10/10 \textit{stratified} train/dev/test split, thus ensuring the same proportion of (frame) labels in each split. Moreover, we generate 4 folds from the dataset, the first used for hyperparmeter search and the remaining three for evaluation.
To this purpose we fine-tune a BERT model\footnote{\url{https://huggingface.co/dbmdz/bert-base-italian-xxl-cased}} \cite{devlin-etal-2019-bert} for Italian and show that the approach allows us to scale with thousands of (unique) labels without a remarkable computational and memory overhead. In this experimental setup we use MaChAmp, v 0.4 beta 2 \cite{van-der-goot-etal-2021-massive}, a toolkit supporting a variety of NLP tasks, including multi-label sequence labeling. We performed hyperparameter search by exploring the space with batch sizes between 8 and 256 and learning rates between 7.5e-4 and 7.5e-3. All other hyperparameters are left unchanged with respect to MaChAmp's default configuration for the multi-sequential task.\footnote{\url{https://github.com/machamp-nlp/machamp/blob/master/docs/multiseq.md}} Overall, 64 configurations have been explored. The best
hyperparameter values we found, according to the performance on the development set, are batch size of 64 and learning rate of 1.5e-3 and the resulting model has been used for the evaluation (Section \ref{sec:results}). The training requires approximately 3.5 hours on an NVIDIA RTX A5000 GPU with 24 GB memory and 8192 CUDA cores. In terms of memory, the maximum requirement is 5 GB RAM.

\section{Evaluation}
\label{sec:results}
In this section, we discuss the quantitative (Section \ref{sec:quantitative}) and qualitative (Section \ref{sec:qualitative}) performance of the multi-label sequence labeling approach on the EvN-ITA dataset.  
%\begin{table}[h]
%    \centering
%    \begin{tabular}{l|rrr}
%    \hline
%    \textbf{Split} & \textbf{Fold 1} & %\textbf{Fold 2} & \textbf{Fold 3} \\
%     \hline
%    Train & 42774 & 42751 & 42662 \\
%    Dev & 5293 & 5405 & 5422 \\
%    Test & 5373 & 5284 & 5356 \\
%    \end{tabular}
%    \caption{Number of sentences per fold/split, used for 3-fold evaluation.}
%    \label{tab:datasplit}
%\end{table}
\subsection{Quantitative results}
\label{sec:quantitative}
Evaluation results are reported in Table \ref{tab:reseventstoken} and Table \ref{tab:reseventsspan} in an aggregated fashion in order to provide the reader with different views on performance. Reported values have been obtained by separately computing the metrics class-wise on each fold, and then averaging the obtained scores. For each of the two groups of labels (frames and frame elements), beside the overall average performance, we provide the average of the \textit{n}-best and \textit{n}-worst performing classes and the average of the \textit{n} most and least frequent classes in the dataset, on the three test sets, with \textit{n} = 40 for frames and \textit{n} = 200 for frame elements.\footnote{In the case of frame elements, given their extremely skewed distribution, resulting in a long tail of rare labels, we also apply a threshold, taking into account only labels occurring at least 5 times in the span-based setting and at least 20 times in the token-based setting.} We also compute the macro average and the weighted macro average of all classes, the latter providing a more realistic view in a context of highly unbalanced label distribution. With a strict F1-score of 0.9 for frames and 0.724 for frame elements, our system shows very promising results for the task. Overall, the results show that, despite being fundamentally token-based, our multi-label sequence tagging approach proves effective also in the identification of (multiple) textual spans in a sentence, scaling well on a dataset involving a very high number of classes. This is further confirmed by the small delta between relaxed and strict performance~values.

\subsection{Error Analysis}
\label{sec:qualitative}
To assess the potential of the proposed approach and the possible inconsistencies, we perform an error analysis on the test sets of the three folds, at token level.
Since in a multi-label setting it is not always possible to establish a univocal correspondence between labels in the gold and predicted sets (given the possibility of multiple assignments on both sides), we proceed as follows: for each token, we filter out from both sets of labels (gold and predicted) the correctly matched labels. Based on this output, we focus on a subset of tokens, those labeled, in both sets, with exactly one label and we use it as an approximation for identifying most common errors.
\begin{table}[]
\resizebox{1\linewidth}{!}{
\begin{tabular}{cc}
    \hline
    \textbf{Gold} & \textbf{Predicted}\\
    \hline
    \textsc{Blaming} & \textsc{Accuse} \\
    \textsc{Hostile\_Encounter} & \textsc{War} \\
    \textsc{Conquering} & \textsc{Occupancy} \\
    \textsc{Accuse} & \textsc{Blaming} \\
    \textsc{Replacing} & \textsc{Take\_Place\_Of} \\
    \textsc{Occupancy} & \textsc{Conquering} \\
    \textsc{Building} & \textsc{Manufacturing} \\
    \textsc{Create\_Artwork} & \textsc{Text\_Creation} \\
    \textsc{Killing} & \textsc{Death} \\
    \textsc{Request} & \textsc{Questioning} \\
\end{tabular}
}
\caption{Top 10 prediction errors between event frames.}
\label{tab:analysisevents}
\end{table}
\begin{table}[]
\resizebox{1\linewidth}{!}{
\begin{tabular}{l}
    \hline
    \textbf{Frame element - Event frame}\\
    \hline
    \textsc{(G) Reason-Blaming}\\
    \textsc{(P) Offense-Accuse}\\
    \hline
    \textsc{(G) Interlocutor2-Conversation} \\
    \textsc{(P) Party2-Negotiation} \\
    \hline
    \textsc{(G) Location-BeingLocated} \\
    \textsc{(P) RelativeLocation-BeingLocated} \\
    \hline
    \textsc{(G) Message-Request} \\
    \textsc{(P) Message-Questioning} \\
    \hline
    \textsc{(G) RelativeLocation-BeingLocated} \\
    \textsc{(P) Location-BeingLocated} \\
    \hline
    \textsc{(G) Message-Answer} \\
    \textsc{(P) Message-Reply} \\
    \hline
    \textsc{(G) Issue-TakingSides} \\
    \textsc{(P) Side-TakingSides} \\
    \hline
    \textsc{(G) Artwork-CreateArtwork} \\
    \textsc{(P) Text-TextCreation} \\
    \hline
    \textsc{(G) Evaluee-Blaming} \\
    \textsc{(P) Accused-Accuse} \\
    \hline
    \textsc{(G) Explanation-Death} \\
    \textsc{(P) Cause-Death} \\
    \hline
\end{tabular}
}
\caption{Top 10 prediction errors between frame elements (G = Gold, P = Predicted).}
\label{tab:analysisroles}
\end{table}
This allows us to focus on specific one-to-one label mismatchings, both for event frames (Table \ref{tab:analysisevents}) and for frame elements (Table \ref{tab:analysisroles}). Considering only event frames, analysis reveals that only 4.8\% of the identified mismatches involves two event labels, while 95.2\% involves a mismatch between an event label and the O-tag (out-of-mention). This ratio becomes more balanced with regard to frame elements (39\% and 61\%, respectively). Also, the impact of errors referred to the IOB schema remains very low, amounting to 1.16\% for event frames and 4.8\% for frame elements.\\ 
Qualitatively, the analysis shows a clear pattern, namely that errors occur in most cases between frames with a high semantic similarity, like \textsc{Blaming}/\textsc{Accuse} or \textsc{Hostile\_Encounter}/\textsc{War}, which in some cases may be difficult to classify even for the human annotator. As for frame elements, errors occur mostly \textit{a)} between the same frame element of two different event frames (for example \textsc{Message-Request} vs. \textsc{Message-Questioning}) or \textit{b)} between frame elements that have a latent semantic correspondence in different frames (\textsc{Interlocutor2-Conversation} vs. \textsc{Party2-Negotiation} or \textsc{Reason-Blaming} vs. \textsc{Offense-Accuse}) or, still, \textit{c)} between semantically close frame elements within the same frame (\textsc{Explanation-Death} vs. \textsc{Cause-Death}). These quite subtle error types further reveal how the multi-label sequence labeling approach is capable of learning cross-frame correspondences of frame elements, an aspect that we plan to further investigate in future work.

\section{Dataset and Model Release}
\label{sec:release}
The EvN-ITA annotated dataset, along with its documentation, is being released upon request\footnote{The dataset can be requested by filling out the form at \url{https://forms.gle/qAgZsf4La9qdzETn6} or by emailing the author at \href{mailto:eventnetita@gmail.com}{eventnetita@gmail.com}.}, under CC-BY-SA 4.0 license\footnote{\url{https://creativecommons.org/licenses/by-sa/4.0/deed.en}}. The model of the frame parser, described in Section \ref{sec:experimental}, is available on Huggingface\footnote{\url{https://huggingface.co/mrovera/eventnet-ita}}.

\section{Conclusion and Future Works}
\label{sec:conclusions}
In this paper we presented EvN-ITA, a large corpus annotated with event frames in Italian, accompained by an efficient multi-label sequential model for Frame Parsing, trained and evaluated on the corpus.
Future work includes extrinsic tests of the resource on new data from different textual genres and the reinforcement of the schema, in view of providing a wider domain coverage and increased adaptability of the model. Moreover, is our plan to employ EvN-ITA as a benchmark to investigate the performance of different methodologies and learning models for Frame Parsing, as well as to explore strategies for multilingual applications.

\section*{Limitations}
The first limitation of this work lies in the unique source of the data, Wikipedia, that, if on the one end guarantees an ample variety of topics and types of events, on the other hand, from the linguistic point of view it sets a constraint on a homogeneous linguistic style. As mentioned above, this will be the focus of our future effort.
Secondly, in case of multiple mentions of the \textit{same} event frame in a given sentence (this case concerns 6\% of the sentences in EvN-ITA), the currently adopted methodology does not support automatic linking of frame elements to the exact frame mention they refer to in the sentence. Future approaches will be geared to take this issue into account.  

\section*{Acknowledgements}
I owe my gratitude to Serena Cristoforetti, for her dedicated contribution to the inter-annotator assessment process, carried out as part of her curricular internship at the Digital Humanities (DH) group. I am equally thankful to Alan Ramponi and Sara Tonelli (DH) and to Enrica Troiano (Vrije Universiteit Amsterdam) for proofreading the manuscript and for their precious and constructive feedback.\\
This work and the effort it represents are dedicated to the memory of Anna Goy.

% Entries for the entire Anthology, followed by custom entries
\bibliography{eacl2023}
\bibliographystyle{acl_natbib}

\clearpage
\appendix

\clearpage
\section{Examples of annotation}
\label{appendix:appendixa}
While the full documentation of EvN-ITA, including annotation guidelines, frame-based descriptions and examples is being released along with the resource, in this section we provide more details about the annotation process.\\
As mentioned in Section \ref{sec:annotation}, in EvN-ITA target parts-of-speech are nouns, verbs and multiword expressions. 
In real-world data, however, beside events expressed in positive, factual form, there are often cases that raise exceptions. In EvN-ITA,
event mentions are annotated regardless of their factuality value, which means that also \textit{negated}, \textit{abstract}, \textit{hypotetical} event mentions must be annotated, as well as those introduced by modals verbs. Conversely, we do not annotate as event mention those lexical units that are used with \textit{metaphorical meaning} or in the form of \textit{rethoric expression}.
In the following, we provide some examples\footnote{For the sake of readability, examples presented in this appendix are annotated with a single layer. Please see the documentation for more data samples.}:
\begin{enumerate}
    \item \textit{negated events}

    \begin{quote}
        [La Federazione russa, l'unico legale stato successore dell'Unione Sovietica \textsc{Payer}], non ha mai riconosciuto le deportazioni degli estoni come un crimine e non ha [\textbf{pagato} \textsc{Pay}] [nessuna riparazione \textsc{Money}] [agli stati coinvolti \textsc{Beneficiary}]. 
    \end{quote}

    \begin{quote}
        La persistente segregazione razziale negli Stati Uniti d‘America in tutto il profondo sud significò che [la maggior parte degli afroamericani \textsc{Member}] non poteva [\textbf{fare parte} \textsc{Membership}] [dei Grand Jury \textsc{Group}] i quali – totalmente composti da bianchi – continuarono ad emanare verdetti discriminatori e palesemente ingiusti. 
    \end{quote}

    \begin{quote}
      [Clary \textsc{Speaker}] però non può [\textbf{ribattere} \textsc{Reply}] perché il suo cellulare squilla:  
    \end{quote}

    \begin{quote}
        
    \end{quote}
    
    \item \textit{hypothetical/possible events}

    \begin{quote}
        Sull'orlo di una [\textbf{guerra} \textsc{War}], la Russia comunicò riluttante a Berlino e Vienna il suo consenso e, abbandonata a sé stessa, il 31 marzo, anche la Serbia si arrese.  
    \end{quote}

    \begin{quote}
        Nel gennaio 2008 hanno iniziato a rincorrersi notizie, via via sempre più insistenti e accreditate, che [ad Albano Laziale \textsc{Place}], [in prossimità della discarica sita in località "Roncigliano" della frazione di Cecchina \textsc{Relative\_Location}], sarebbe stato [\textbf{realizzato} \textsc{Building}] [un inceneritore \textsc{Created\_Entity}] per smaltire i rifiuti, in vista dell'imminente chiusura della discarica di "Roncigliano" e di quella di Malagrotta a Roma.  
    \end{quote}

    \begin{quote}
        Il 22 novembre 1961 la polizia perquisì, senza risultato, il suo appartamento in cerca di [una pistola con cui \textsc{Means}] [Pasolini] avrebbe [\textbf{rapinato} \textsc{Robbery}], [il 18 sera \textsc{Time}], [un distributore di benzina \textsc{Source}] [di San Felice Circeo \textsc{Place}].
    \end{quote}

    \begin{quote}
        La bomba non esplose, altrimenti [la detonazione \textsc{Cause}] avrebbe potuto effettivamente [\textbf{distruggere} \textsc{Destroying}] [la nave \textsc{Patient}].        
    \end{quote}

    \begin{quote}
        Il riluttante generale Raffaele Cadorna, per evitare che [Mussolini \textsc{Captive}] [\textbf{cadesse nelle mani} \textsc{Taking\_Captive}] [degli Alleati \textsc{Captor}], rilasciò il salvacondotto necessario;
    \end{quote}
    
    \item \textit{events introduced by modals}

    \begin{quote}
        Verosimilmente, si può presumere che [egli \textsc{Mover}] dovette [\textbf{allontanarsi} \textsc{Quitting\_a\_Place} [da Augsburg \textsc{Source}] [in quanto cattolico \textsc{Explanation}], dal momento c
        he, con quanto sancito dalla pace di Augusta, era in vigore il principio del "cuius regio, eius religio".
    \end{quote}

    \begin{quote}
        Se l'indagine dimostrerà che sono stati commessi crimini di guerra, [i responsabili \textsc{Defendant}] dovranno essere trovati e [\textbf{processati} \textsc{Trial}] [conformemente alle norme in vigore \textsc{Binding\_Principle}].
    \end{quote}
\end{enumerate}
Conversely, we do \textit{not} annotate as event mention occurrences used with \textit{metaphorical meaning} or in the form of \textit{rethoric expression}:

\begin{quote}
Questa è [\textbf{la domanda} Ø] che Malone, Russo e Montague hanno cominciato a [\textbf{porsi} Ø] una volta esaurita la spinta idealistica dei primi anni di lavoro e la loro risposta è stata:
\end{quote}

\begin{quote}
Signor Presidente, il risultato delle elezioni in Israele ha [\textbf{fornito la risposta} Ø] della popolazione israeliana. 
\end{quote}

\begin{quote}
    Il visitatore / studioso poteva [\textbf{intraprendere} Ø] così [\textbf{un viaggio} Ø] dal microcosmo (la chimica), attraverso gli elementi primi della natura, al macrocosmo (l'astronomia) nel torrino che concludeva il percorso.
\end{quote}
\begin{quote}
    Sostiene inoltre che, nonostante i problemi della filosofia della scienza e della ragione in generale, le "questioni morali" avranno [\textbf{risposte} Ø] oggettivamente giuste e sbagliate suffragate da fatti empirici su ciò che induce la gente a star bene e prosperare.
\end{quote}

Finally, as mentioned in Section \ref{sec:annotation}, in order to increase robustness, EvN-ITA contains many negative examples. Given a lexical unit, a negative example is an occurrence of the lexical unit which denotes a meaning not covered by the current schema. Negative examples are meant to improve the classifier's ability to work in an open-world setting and to generalize to extrinsic/unseen data.

Lexical unit: \textit{istituire}

Positive example:
\begin{quote}
    La richiesta venne accolta e il papa diede l'autorizzazione a [\textbf{istituire} \textsc{Create\_Social\_Entity}] [in Inghilterra \textsc{Place}] [un tribunale ecclesiastico \textsc{Created\_Entity} [per esaminare attentamente il caso \textsc{Purpose}], …. 
\end{quote}

Negative example:
\begin{quote}
    venne così [\textbf{istituito} Ø], nel 46 a.C., il calendario giuliano.
\end{quote}

\clearpage

\section{Association between event frames}
\label{appendix:appendixb}
In this section we present numerical evidence of association between events, mentioned in Section \ref{sec:experimental}. As stated above, patterns of association between frames can be identified by computing their co-occurrence. We choose 5 event frames and list the first 5 most related event frames and the 5 most unrelated frames, using Pointwise Mutual Information (PMI).
\begin{table}[h!]
    \centering
    \begin{tabular}{l l}
    \multicolumn{2}{c}{Target: \textsc{Invest}} \\
    \hline
    Related frames & PMI\\
    \hline
    \textsc{Buy} & 2.87\\
    \textsc{Growth on a scale} & 2.37\\
    \textsc{Building} & 2.22\\
    \textsc{Sell} & 2.20\\
    \textsc{Manufacturing} & 1.79\\
    ... & \\
    \textsc{Communication} & -0.87\\
    \textsc{Death} & -0.89\\
    \textsc{Conquering} & -0.93\\
    \textsc{Statement} & -1.04\\
    \textsc{Attack} & -1.38\\
    \end{tabular}
    \caption{Correlation with the \textsc{Invest} frame.}
    \label{tab:correlation1}
\end{table}

\begin{table}[h]
    \centering
    \begin{tabular}{l l}
    \multicolumn{2}{c}{Target: \textsc{Arriving}} \\
    \hline
    Related frames & PMI\\
    \hline
    \textsc{Departing} & 1.94\\
    \textsc{Encounter} & 1.86\\
    \textsc{Move away} & 1.82\\
    \textsc{Remain in place} & 1.78\\
    \textsc{Motion downwards} & 1.75\\
    ... & \\
    \textsc{Being married} & -1.55\\
    \textsc{Decrease on a scale} & -1.58\\
    \textsc{Acquittal} & -1.59\\
    \textsc{Earthquake} & -1.62\\
    \textsc{Taking sides} & -1.79\\
    \end{tabular}
    \caption{Correlation with the \textsc{Arriving} frame.}
    \label{tab:correlation2}
\end{table}

\begin{table}[]
    \centering
    \begin{tabular}{l l}
    \multicolumn{2}{c}{Target: \textsc{Create artwork}} \\
    \hline
    Related frames & PMI\\
    \hline
    \textsc{Performing arts} & 2.18\\
    \textsc{Assign task} & 1.89\\
    \textsc{Temporal origin} & 1.73\\
    \textsc{Publishing} & 1.72\\
    \textsc{Being located} & 1.67\\
    ... & \\
    \textsc{Purpose} & -1.49\\
    \textsc{Robbery} & -1.50\\
    \textsc{Appointing} & -1.51\\
    \textsc{Process end} & -1.59\\
    \textsc{War} & -1.70\\
    \end{tabular}
    \caption{Correlation with the \textsc{Create artwork} frame.}
    \label{tab:correlation3}
\end{table}

\begin{table}[h]
    \centering
    \begin{tabular}{l l}
    \multicolumn{2}{c}{Target: \textsc{Trial}} \\
    \hline
    Related frames & PMI\\
    \hline
    \textsc{Acquittal} & 3.81\\
    \textsc{Sentencing} & 3.59\\
    \textsc{Verdict} & 3.12\\
    \textsc{Accuse} & 2.99\\
    \textsc{Execution} & 2.80\\
    ... & \\
    \textsc{Fleeing} & -1.29\\
    \textsc{Being located} & -1.36\\
    \textsc{Create artwork} & -1.37\\
    \textsc{Beat opponent} & -1.39\\
    \textsc{Agreement} & -1.41\\
    \end{tabular}
    \caption{Correlation with the \textsc{Trial} frame.}
    \label{tab:correlation4}
\end{table}

\begin{table}[h]
    \centering
    \begin{tabular}{l l}
    \multicolumn{2}{c}{Target: \textsc{Communication}} \\
    \hline
    Related frames & PMI\\
    \hline
    \textsc{Contacting} & 2.64\\
    \textsc{Questioning} & 2.09\\
    \textsc{Encounter} & 2.04\\
    \textsc{Awareness} & 2.02\\
    \textsc{Giving} & 1.88\\
    ... & \\
    \textsc{Event ordering} & -1.22\\
    \textsc{Counterattack} & -1.24\\
    \textsc{Appointing election} & -1.25\\
    \textsc{Suppressing} & -1.25\\
    \textsc{Take place of} & -1.53\\
    \end{tabular}
    \caption{Correlation with the \textsc{Communication} frame.}
    \label{tab:correlation5}
\end{table}

\end{document}